\documentclass[11pt,twoside]{article}

\usepackage{asp2014}

\aspSuppressVolSlug
\resetcounters

\begin{document}

\title{Improving astroBERT using Semantic Textual Similarity}

\author{Felix~Grezes$^1$,
        Thomas~Allen, 
        Sergi~Blanco-Cuaresma,
        Alberto~Accomazzi,
        Michael~J.~Kurtz,
        Golnaz~Shapurian,
        Edwin~Henneken,
        Carolyn~S.~Grant,
        Donna~M.~Thompson,
        Timothy~W.~Hostetler,
        Matthew~R.~Templeton,
        Kelly~E.~Lockhart,
        Shinyi~Chen,
        Jennifer~Koch,
        Taylor~Jacovich,
        and Pavlos~Protopapas.
} 
\affil{$^1$Harvard-Smithsonian Center for Astrophysics, Cambridge, MA, USA; \email{felix.grezes@cfa.harvard.edu}}

\paperauthor{Felix~Grezes}{felix.grezes@cfa.harvard.edu}{0000-0001-8714-7774}{Harvard-Smithsonian Center for Astrophysics}{HEAD}{Cambridge}{MA}{02138}{USA}
\paperauthor{Thomas~Allen}{thomas.allen@cfa.harvard.edu}{0000-0002-5532-4809}{Harvard-Smithsonian Center for Astrophysics}{HEAD}{Cambridge}{MA}{02138}{USA}
\paperauthor{Golnaz~Shapurian}{gshapurian@cfa.harvard.edu}{0000-0001-9759-9811}{Harvard-Smithsonian Center for Astrophysics}{HEAD}{Cambridge}{MA}{02138}{USA}
\paperauthor{Sergi~Blanco-Cuaresma}{sblancocuaresma@cfa.harvard.edu}{0000-0002-1584-0171}{Harvard-Smithsonian Center for Astrophysics}{HEAD}{Cambridge}{MA}{02138}{USA}
\paperauthor{Alberto~Accomazzi}{aaccomazzi@cfa.harvard.edu}{0000-0002-4110-3511}{Harvard-Smithsonian Center for Astrophysics}{HEAD}{Cambridge}{MA}{02138}{USA}
\paperauthor{Michael~J.~Kurtz}{kurtz@cfa.harvard.edu}{0000-0002-6949-0090}{Harvard-Smithsonian Center for Astrophysics}{HEAD}{Cambridge}{MA}{02138}{USA}
\paperauthor{Edwin~A.~Henneken}{ehenneken@cfa.harvard.edu}{0000-0003-4264-2450}{Harvard-Smithsonian Center for Astrophysics}{HEAD}{Cambridge}{MA}{02138}{USA}
\paperauthor{Carolyn~S.~Grant}{cgrant@cfa.harvard.edu}{0000-0003-4424-7366}{Harvard-Smithsonian Center for Astrophysics}{HEAD}{Cambridge}{MA}{02138}{USA}
\paperauthor{Donna~M.~Thompson}{dthompson@cfa.harvard.edu}{0000-0001-6870-2365}{Harvard-Smithsonian Center for Astrophysics}{HEAD}{Cambridge}{MA}{02138}{USA}
\paperauthor{Timothy~W.~Hostetler}{thostetler@cfa.harvard.edu}{0000-0001-9238-3667}{Harvard-Smithsonian Center for Astrophysics}{HEAD}{Cambridge}{MA}{02138}{USA}
\paperauthor{Matthew~R.~Templeton}{matthew.templeton@cfa.harvard.edu}{0000-0003-1918-0622}{Harvard-Smithsonian Center for Astrophysics}{HEAD}{Cambridge}{MA}{02138}{USA}
\paperauthor{Kelly~E.~Lockhart}{kelly.lockhart@cfa.harvard.edu}{0000-0002-8130-1440}{Harvard-Smithsonian Center for Astrophysics}{HEAD}{Cambridge}{MA}{02138}{USA}
\paperauthor{Shinyi~Chen}{shinyi.chen@cfa.harvard.edu}{0000-0002-7641-7051}{Harvard-Smithsonian Center for Astrophysics}{HEAD}{Cambridge}{MA}{02138}{USA}
\paperauthor{Jennifer~Koch}{jennifer.koch@cfa.harvard.edu}{0000-0001-9231-8689}{Harvard-Smithsonian Center for Astrophysics}{HEAD}{Cambridge}{MA}{02138}{USA}
\paperauthor{Taylor~Jacovich}{taylor.jacovich@cfa.harvard.edu}{000-0003-0226-0343}{Harvard-Smithsonian Center for Astrophysics}{HEAD}{Cambridge}{MA}{02138}{USA}
\paperauthor{Pavlos~Protopapas}{pprotopapas@g.harvard.edu}{0000-0002-8178-8463}{Harvard}{SEAS}{Cambridge}{MA}{02138}{USA}


\begin{abstract}
The NASA Astrophysics Data System (ADS) is an essential tool for researchers that allows them to explore the astronomy and astrophysics scientific literature, but it has yet to exploit recent advances in natural language processing. At ADASS 2021, we introduced astroBERT, a machine learning language model tailored to the text used in astronomy papers in ADS.\\
In this work we:\\
\begin{minipage}{0.85\columnwidth}
\begin{enumerate}
    \item announce the first public release of the astroBERT language model;
    \item show how astroBERT improves over existing public language models on astrophysics specific tasks;
    \item and detail how ADS plans to harness the unique structure of scientific papers, the citation graph and citation context, to further improve astroBERT.
\end{enumerate} 
\end{minipage}

\end{abstract}



\section{Introduction}
The NASA Astrophysics Data System\footnote{\url{https://ui.adsabs.harvard.edu}} (ADS) has been an essential tool for astrophysicists since it was announced at ADASS II by \citet{1993ASPC...52..132K}. With a  bibliographic collection of over 15 million records, discoverability (of authors, missions, telescopes, etc ...) can be challenging even for the most seasoned ADS user, despite the help of the existing search tools within ADS (filters, Boolean operators, citation graph and more).
A particular difficulty stems from the ambiguity of language e.g. tagging papers with results from the "Planck mission" currently requires manually differentiation from the person, the institute, the constant with the same name. Furthermore new entities are constantly being named, and need to be searchable.  
Introduced at ADASS XXXI by \citet{2021arXiv211200590G}, we publicly release an improved version of \textbf{astroBERT}, a language model tailored to astrophysics.
We show that astroBERT outperforms the existing public languages BERT \citep{2018arXiv181004805D} and SciBERT \citep{2019arXiv190310676B} on the task of detecting entities of interest in astrophysics literature. Finally we detail our plans to improve astroBERT using information unique to scientific literature, the citation graph.

\section{astroBERT Pretraining}
We trained the base astroBERT language model on $\sim$400K recent astrophysics papers from ADS comprised in total of $\sim$4B tokens. The pretraining tasks were Masked Language Modeling (MLM) and Next Sentence Prediction (NSP). Pretraining lasted for 40 epochs over $\sim$50 days. 
All computation was done with two NVIDIA V100 GPUs.

\section{Detecting Entities}
To properly compare astroBERT to the existing language models BERT and SciBERT, we finetuned all three models on the Detecting Entities in the Astrophysics Literature (DEAL)\footnote{\url{https://ui.adsabs.harvard.edu/WIESP/2022/SharedTasks} \label{DEAL} } task \citep{grezes-etal-2022-overview-deal}, which consists of identifying and labeling Named Entities in a dataset composed by full-text fragments and acknowledgements from the astrophysics literature. The 32 types of entities are of particular interest to astrophysicists: celestial objects, observational techniques, organizations and persons, etc\dots Full list and definitions with examples available in footnote \ref{DEAL}. Finetuning took $\sim$12 hours.


\section{Public Release}
We release the astroBERT model to the public via the popular Huggingface hub, along with its associated WordPiece tokenizer and tutorials. The astrophysics research community can start using astroBERT in minutes, on predefined tasks such as detecting named entities in astrophysics literature or finetuning for their own custom tasks.
\begin{center}
     \includegraphics[height=0.4cm]{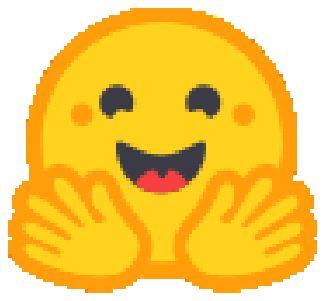} \url{https://huggingface.co/adsabs/astroBERT} 
\end{center}
\articlefigure[width=1.\textwidth]{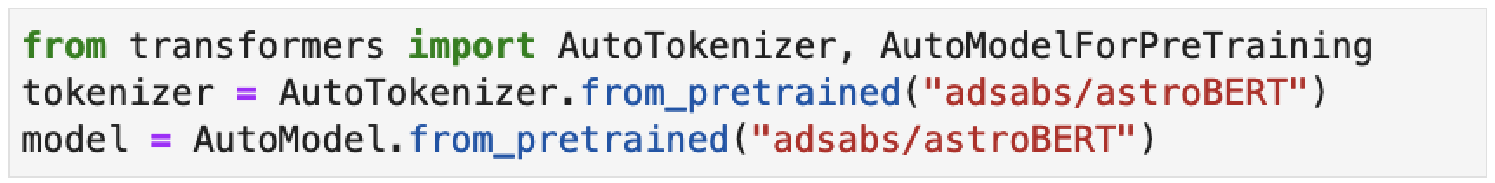}{fig-code}{All you need to get you started with astroBERT.}%

\section{Results: Comparison to BERT and SciBERT}
Table \ref{tab:metrics} shows how the current astroBERT outperforms BERT, sciBERT and the version of astroBERT that was used as the baseline for DEAL. The models were compared using  Matthews correlation coefficient (MCC), which takes into account every value in the confusion matrix and is generally regarded a balanced measure, along with the standard micro-averaged F-1, precision, and recall scores. Additionally, we included a model making random predictions based on label frequency for comparison.

\begin{table}[!ht]
\caption{astroBERT outperforms other language models on astrophysics data.}
\smallskip
\begin{center}
\resizebox{1.\columnwidth}{!}{
\begin{tabular}{|c|c|c|c|c|c|c|}
\hline
 &
  Metric $\Longrightarrow$  &
  MCC &
  overall  &
  overall  &
  overall  &
  overall  \\ 
Model  &
  Split $\Downarrow$ &
   &
   F1 score &
   precision\ &
   recall &
   accuracy \\ \hline \hline 
  
 &
  train &
  0.1037 &
  0.0170 &
  0.0122 &
  0.0278 &
  0.7146 \\ \cline{2-7} 
 Random &
  val &
  0.1083 &
  0.0166 &
  0.0119 &
  0.0273 &
  0.7059 \\ \cline{2-7} 
 &
  test &
  0.1057 &
  0.0162 &
  0.0116 &
  0.0269 &
  0.6876 \\ \hline \hline
 &
  train &
  0.7542 &
  0.4920 &
  0.4995 &
  0.4848 &
  0.9256 \\ \cline{2-7} 
 BERT &
  val &
  0.7405 &
  0.4739 &
  0.4780 &
  0.4698 &
  0.9188 \\ \cline{2-7} 
 &
  test &
  0.7229 &
  0.4513 &
  0.4622 &
  0.4409 &
  0.9094 \\ \hline \hline
 &
  train &
  0.8159 &
  0.5867 &
  0.5753 &
  0.5986 &
  0.9430 \\ \cline{2-7} 
 SciBERT &
  val &
  0.8019 &
  0.5601 &
  0.5463 &
  0.5745 &
  0.9366 \\ \cline{2-7} 
 &
  test &
  0.7844 &
  0.5355 &
  0.5313 &
  0.5398 &
  0.9280 \\ \hline \hline
 &
  train &
  0.8296 &
  0.6138 &
  0.5889 &
  0.6409 &
  0.9468 \\ \cline{2-7} 
astroBERT &
  val &
  0.8104 &
  0.5779 &
  0.5508 &
  0.6077 &
  0.9389 \\ \cline{2-7} 
(WIESP) &
  test &
  0.7939 &
  0.5561 &
  0.5387 &
  0.5746 &
  0.9308 \\ \hline \hline
 &
  train &
  0.8250 &
  0.5995 &
  0.5701 &
  0.6319 &
  0.9442 \\ \cline{2-7} 
\textbf{astroBERT} &
  val &
  0.8194 &
  0.5907 &
  0.5575 &
  0.6282 &
  0.9405 \\ \cline{2-7} 
(public release) &
  test &
  \textbf{0.8302} &
  \textbf{0.6093} &
  \textbf{0.5846} &
  \textbf{0.6362} &
  \textbf{0.9418} \\ \hline
\end{tabular}
}
\end{center}
\label{tab:metrics}
\end{table}

Figure \ref{fig-astrosci} compares the performances of astroBERT to that of SciBERT over each label along both precision and recall. Larger circles indicate a higher number of occurrences of that label in the data, while the color indicates if the label is mostly present in the fulltext (red) or the acknowledgment (blue) section of the paper. This tells us where astroBERT does well, for example on Observational Techniques (ObT) and Database (DaB), and which areas of to focus on to improve astroBERT further, URLs which can be captured using regular expressions, and which labels that will benefit the most from additional data, for example Event (Eve) and Identifier (Ide) (see footnote \ref{DEAL} the meaning of the abbreviations). 

\articlefigure[width=1.\textwidth]{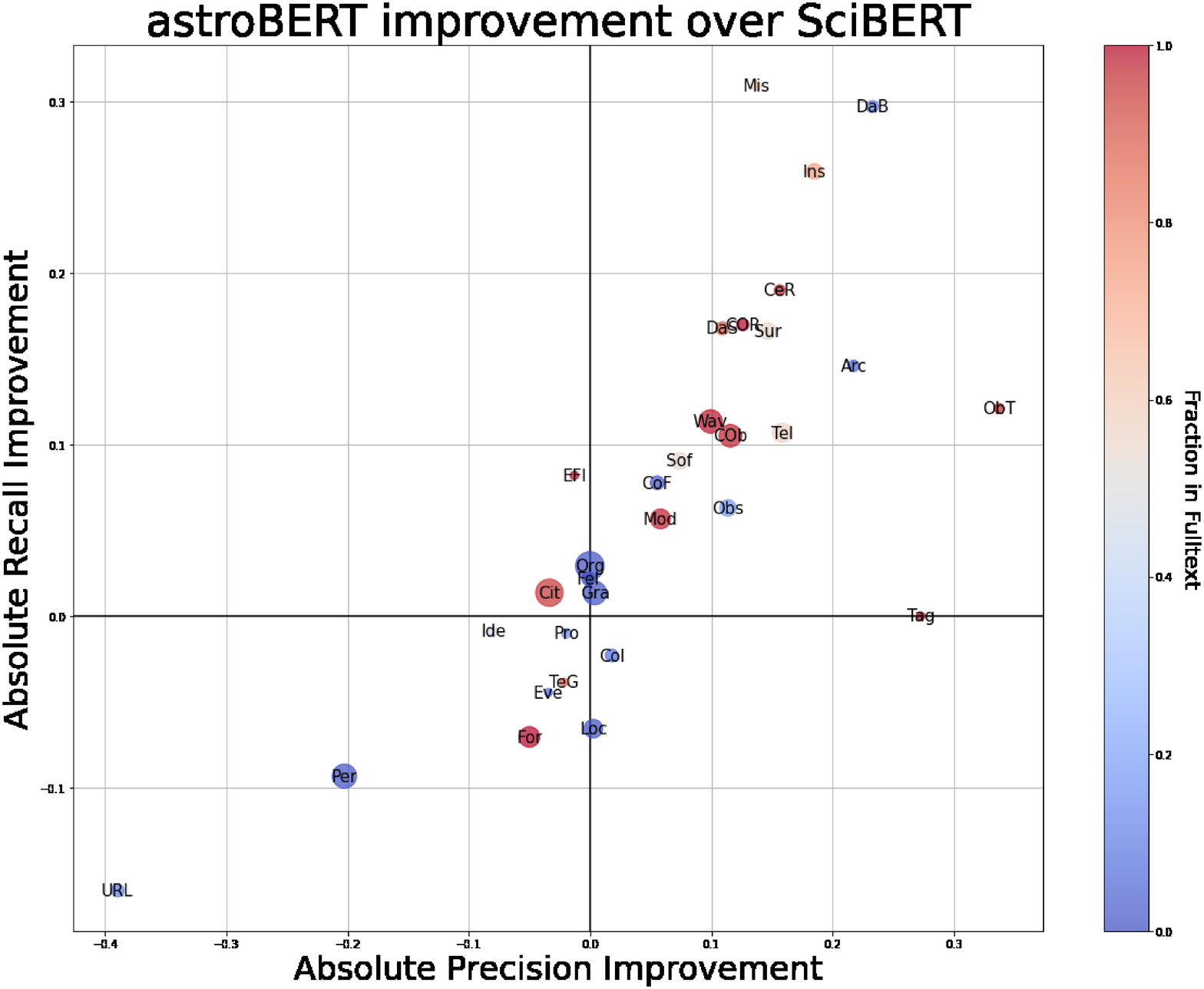}{fig-astrosci}{Detailed comparison between astroBERT and SciBERT.}

\section{Harnessing Semantic Similarity}
While the current version of astroBERT uses MLM and NSP as unsupervised training tasks, it does not exploit the information unique to scientific papers, i.e. the citation graph and the context of any given citation. Based on the works of \citet{2019arXiv190810084R} on Sentence-BERT, we will train astroBERT on semantically similar sentences extracted from the context of citations i.e. abstracts from the cited paper paired with the paragraph preceding the citation. This pairing should be produce sentences embeddings more useful to astrophysics tasks than NSP or the weakly labeled sentence pairings used by Reimers et al. from  \citet{agirre-etal-2013-sem}.

\newpage
\bibliography{P45}  


\end{document}